# Understanding the Artificial Intelligence Clinician and optimal treatment strategies for sepsis in intensive care

Matthieu Komorowski[1,2], Leo A. Celi[3,4], Omar Badawi[3,5,6], Anthony C. Gordon[1,*] & A. Aldo Faisal[2,8,9,10,11,*]

[*]Corresponding authors: Anthony C. Gordon (anthony.gordon@imperial.ac.uk) & A. Aldo Faisal(aldo.faisal@imperial.ac.uk)

1. Department of Surgery and Cancer, Imperial College London, London, UK
2. Department of Bioengineering, Imperial College London, London, UK
3. Laboratory of Computational Physiology, Harvard–MIT Division of Health Sciences & Technology, Cambridge, MA, USA
4. Beth Israel Deaconess Medical Center, Boston, MA, USA
5. Patient Care Analytic, Philips Healthcare, Baltimore, MD, USA
6. Department of Pharmacy Practice and Science, University of Maryland, School of Pharmacy, Baltimore(MD),USA
7. Department of Surgery and Cancer, Imperial College London, London, UK
8. Centre in AI for Healthcare, Imperial College London, London, UK
9. Department of Computing, Imperial College London, London, UK
10. MRC London Institute of Medical Sciences, London, UK
11. Behaviour Analytics Lab, Data Science Institute, London, UK

In this document, we explore in more detail our published work (Komorowski, Celi, Badawi, Gordon, & Faisal, 2018) for the benefit of the AI in Healthcare research community. In the above paper, we developed the AI Clinician system, which demonstrated how reinforcement learning could be used to make useful recommendations towards optimal treatment decisions from intensive care data. Since publication a number of authors have reviewed our work (e.g. Abbasi, 2018; Bos, Azoulay, & Martin-Loeches, 2019; Saria, 2018). Given the difference of our framework to previous work, the fact that we are bridging two very different academic communities (intensive care and machine learning) and that our work has impact on a number of other areas with more traditional computer-based approaches (biosignal processing and control, biomedical engineering), we are providing here additional details on our recent publication. We acknowledge the online comments by Jeter et al (https://arxiv.org/abs/1902.03271). The sections of the present document are structured so as to address some of their questions. For clarity, we label figures from our main Nature Medicine publication as "*M*", figures from Jeter et al.'s arXiv paper as "*J*" and figures from our response here as "*R*".

Jeter et al. state "the only possible response we can afford is a more aggressive and open dialogue". We are pleased to clarify the points they raise, engage in a scientific discussion about the methods and results of our research, all of which were and are fully open and transparent. We feel as academics such a dialogue should be professional and robust, not aggressive and sensational. Open dialogue should also mean that all parties' conflicts of interest are declared, which we disclosed on our part in the Nature Medicine publication and again here.

# General comments on the safety of the AI Clinician

In our academic research paper, we are demonstrating the proof-of-concept of how reinforcement learning may be a useful tool to improve difficult clinical decision making in sepsis. It is therefore important to understand what we show in this paper and what we do not show nor claim. The AI clinician as a working medical device or more precisely speaking a clinical decision support system is nowhere near ready as a medical device ready to be used in closed-loop, directly recommending drug dosages to patients. We are surprised that Jeter et al. would confound the basic research presented in our paper with that of a finished medical product given our explicit signposting to that end.

Crucially, as we explained our work was only evaluated retrospectively on observational data, and not in closed loop interactively with patients. Thus, what we present is not a medical device that is ready for clinical use. We agree with Jeter et al. that patient safety is of utmost priority and as such, we have always maintained that the motivation to develop a model that could be used as part of a decision support system to aid fellow clinicians in the care of patients who have sepsis. Therefore, our stated objective in the Nature Medicine paper was not to publish a medical device that will implicitly improve outcomes from sepsis across all hospitals and forever, but demonstrate instead the principles of how observational data can be harnessed for reinforcement learning applied to sepsis treatment. In fact, in January 2019 we were part of a consortium that published "Guidelines for reinforcement learning in healthcare" (Gottesman et al., 2019) by which such an approach observational data can inform the deployment of reinforcement learning in healthcare in a safe, risk-conscious manner.

Our aim was to demonstrate how day-to-day data as captured by the digitalization of care in situations where best practice is not established may assist with identifying optimal decisions. We believe that AI clinician approach can improve the huge variability in practice in sepsis resuscitation and narrow down the range of possible actions to a more rational set of choices. The AI Clinician was designed with patient safety in mind. For example, we have limited the set of decisions available to the AI clinician only to decisions made relatively frequently by human clinicians. As such, we turned the AI agent from a free-roaming entity exploring an *in silico* environment into an agent suggesting best decisions among the ones taken frequently by clinicians.

We have evidence of this. *Figure R1* shows the distribution of the fraction of patients who actually followed the optimal action, in all the 750 states of the model. It shows that over 2/3 of the time, the AI clinician suggests choosing an action that was taken at least 5% of the time in the training data. Actions associated with rare or unseen transitions are not available paths for the AI agent.

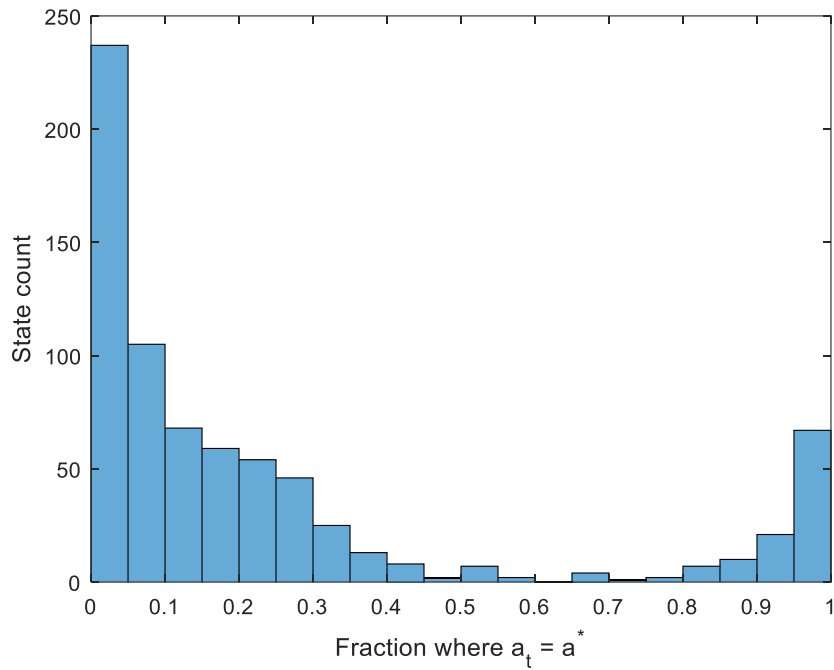

*Figure R1: Fraction of patients who followed the optimal action, in all the 750 states of the model. In more than two thirds of the states (513 states out of 750), the suggested optimal action was taken by more than 5% of the clinicians in the training dataset. The AI clinician does not suggest rare decisions that were nearly never taken in real life. $a_t = a^*$ refers to instances where an actual chosen action at time t is the optimal action in a given state.*

## Goodness of fit of the transition matrix

Jeter et al. ask for more evidence of the goodness of fit of the transition matrix. This is what we did: we used Monte Carlo simulations to assess this. Specifically, we used the model to generate virtual trajectories, starting from randomly sampled initial states, following the learned transition matrix and the clinicians' policy (*Figure R2*) until reaching an absorbing state (death or successful discharge). We generated 1,000 batches of 2,500 virtual trajectories (for a total of 2.5 million virtual trajectories) to produce estimates of the distribution. The results of this model-based simulation are very accurate: the predicted mortality is 22.47% (SD 0.86%) to be compared to an actual mortality of 22.5% and the average predicted length of trajectories is 14.51 time-steps (SD 0.23) whereas it is 14.42 time-steps (SD 3.75) in the data. Length restriction did not allow us to include these secondary results in the paper.

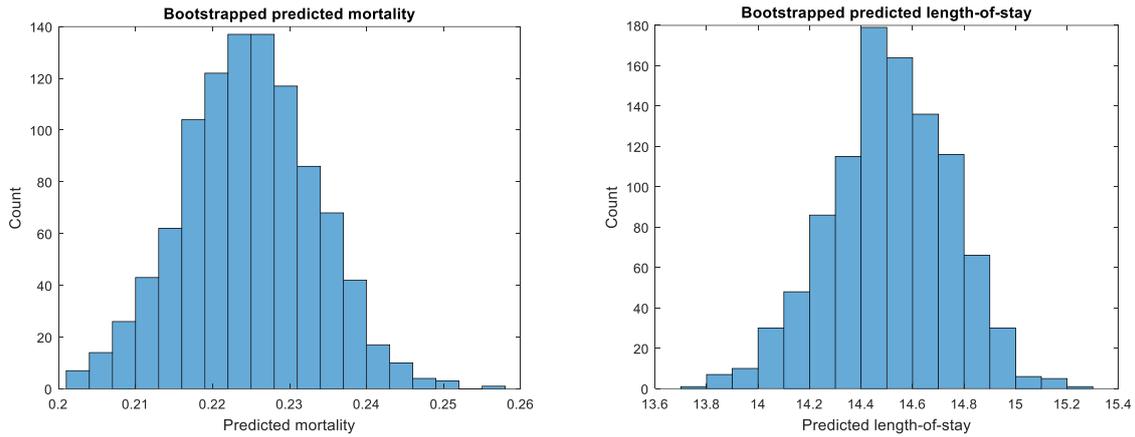

*Figure R2: Predicting actual patient mortality and length of trajectories with Monte Carlo simulations.*

# High value of zero-drug policy

Regarding *figure J2*, Jeter et al. rightly describe that "for most of the realizations, the zero-drug policy outperforms the clinician policy." This statement overlooks the way WIS estimation operates by effectively weighing data so as to only consider healthy patients unlike the clinician or AI policy that operated on all patients. Recall, that WIS estimation of a zero-drug policy will discard any actual data where patients did receive any drug, e.g. it only evaluates the healthiest patients. Jeter et al. argue that the same phenomenon is also happening with the AI policy, but it is obviously not the case. The zero-drug and the AI policy are highly different in terms of actions. In fact, the clinicians' policy is closer to the zero-drug policy than the AI policy, but it has the lowest value among the three according to *figure J2*. The distribution of actions for the AI clinician (*Figures M3b and M3c*) show that it gives higher doses of vasopressors on average and a more restrictive strategy for IV fluids. If the bias towards healthy patients on no treatment were prominent, then the AI clinician would predominantly recommend giving no vasopressor and no fluids (action 1). This is clearly not the case.

# Reproducing the AI Clinician

Reproduction is essential to research. That is why we have not only described the mathematical methods used but also made all AI clinician code available (as is Nature Medicine publication policy) but also pre-processing pipelines for the data. The databases and the raw data itself, as we explained, cannot be directly made available by us as it is governed by access processes of the organisations that oversee and manage this patient data. We explain how data access can be requested, and details on the pipeline to extract our specific dataset from the databases. Upon publication in November 2018, we were contacted by a very large number of interested individuals (including Dr Jeter) asking for the code to try out our system. To better handle the scale of interest, we set up a public website and informed all those that contacted that we would be in touch when this was completed. The AI Clinician is now part of the official Imperial College university website (http://www.imperial.ac.uk/artificial-intelligence/research/healthcare/ai-clinician/) coming online in January 2019. We then contacted again all individuals (including Dr Jeter) the link to the public website containing our AI Clinician code and instructions on use.

Jeter et al. attempt to reproduce the AI Clinician results but use a small fraction of the original dataset ("We utilized data from 5,366 septic patients from the MIMIC III dataset in which every patient received both vasopressors and fluids") and their own model (instead of ours). This precludes drawing any conclusions about the reproducibility of our results from their results. We would contend that including only patients who received both fluids and vasopressors has limited clinical utility. This is because some of the hardest decisions in medical practice are knowing when it is the right time to start a therapy or not. Therefore, it is vital to include a broad population of patients available who have sepsis, as we have argued and done in our original work.

Moreover, we employed datasets that can be publicly accessed, and we have published our code and even the database queries because we would like to invite others to build on what we have done. We are delighted to see that well over a dozen preprints on arXiv have currently (March 2019) been published that use our sepsis specific dataset to further develop reinforcement learning in healthcare research.

## Single trajectory

In *figure J5*, Jeter et al. describe a time series of a patient's mean arterial pressure (MAP), along with the given doses of drugs, and the doses suggested by their agent. No inference can be drawn about the behaviour of the AI Clinician from a single cherry-picked trajectory shown by Jeter et al. First of all, it represents the result of a very different model from the AI clinician. Moreover, even if the model where the same, we do not believe in "proof-by-example", as outliers are always expected in living systems. From a clinical perspective we also question the specific example shown: What was the clinical context of this patient? Clinicians who treated this patient tolerated a MAP of 50 mmHg with little to no increase in vasopressor dose for more than 10 hours, at face value a very unusual decision. External factors most probably explain this peculiar profile and it is possible that this patient should have been excluded from their training data. To be clear, we are not suggesting that we have excluded such patients from our dataset, as we have included tens of thousands of patients. Irrespective of the training data used, this is another reason why we stressed in our original work, by prospective testing of the AI Clinician is necessary before it can be used clinically.

## Importance sampling

As correctly explained by Jeter et al., the basic principle of importance sampling is that segments of trajectories where the two policies (the behaviour policy and the evaluation policy) agree are taken into account to quantify the value of the policy evaluated. When the two policies differ, little inference can be drawn, which is true regardless of which agent (AI or clinician) took the best decision. This holds for both the situations where the AI clinician "acted wrongly", or the human clinicians "acted wrongly" and there is no indication that the AI policy is favouring the least sick patients or impossible transitions from severe states to healthy states.

In addition, we fully acknowledge that the off-policy evaluation is a very difficult problem and an active area of research. Some of the considerations and trade-offs in off-policy evaluations, as well as their illustration using the very dataset we used for our original model have been recently published as papers or preprints (Li, Komorowski, & Faisal, 2018; Liu et al., 2018;

Peng et al., 2019; Raghu et al., 2018). Either way, we should highlight that as stated in our methods, "WIS may be a biased although consistent policy estimator, so the bootstrap confidence interval may also be biased, even though the literature suggests that consistency is a more desirable property than unbiasedness (Hanna, Stone, & Niekum, 2016; Jiang & Li, 2015; Precup, Sutton, & Singh, 2000). It is accepted that bootstrapping produces accurate confidence intervals with less data than exact HCOPE methods, and is safe enough in high risk applications such as healthcare (Hanna et al., 2016; Thomas, Theocharous, & Ghavamzadeh, 2015)." The HCOPE algorithm we used arguably represents the current state-of-the-art, and we are open to suggestions to explore other methods that may eventually prove to be more appropriate.

## Short-term or long-term reward?

Regarding short-term resuscitation goals, we agree with Jeter et al. that avoiding hypotension is an important part of sepsis management. However, there is still uncertainty about what blood pressure is optimal and the use of surrogate markers is not without limitations. Previous randomised controlled trials have shown no difference in mortality rates between different blood pressure targets in sepsis. Jeter et al. highlight that 65 mmHg is the current recommended target but there are ongoing trials testing if lower blood pressure targets may be better for some septic patients (E.g. the 65 trial: http://www.isrctn.com/ISRCTN10580502). It is likely that patients need individualised targets and that these might vary over time. There is also plenty of evidence that targeting or improving short-term physiological rewards (blood pressure (Asfar et al., 2014; Lamontagne et al., 2016), urine output (Myburgh et al., 2012), oxygenation (ARDSNet, 2000)) can ultimately lead to worse long-term survival. We have therefore selected longer-term survival (90 days) as our reward signal as this is what matters to patients. The reinforcement learning approach allows us to assess and operate towards longer-term outcomes from a series of decisions, instead of just single action outcomes as in many causal modelling applications.

In addition, we would argue that human clinicians often perform quite poorly when trying to balance short-term and long-term goals. For example, it is common to try and improve blood pressure and urine output with (often excessive) fluid boluses, when this may impair long term organ function and survival. Most experienced clinicians will frequently remind their junior colleagues that "making the numbers look good" may not be the best treatment strategy for patients.

## Choice of a 4h time resolution

Jeter et al. propose alternatives to our approach such as fast feedback goal-directed loop, for example where vasopressors or fluids would be administered in response to a reduction in MAP. This is an active area of research, with several prominent teams offering solutions (Joosten et al., 2018; Rinehart et al., 2015).

While these tools may be successful at restoring blood pressure, they merely replicate what clinicians have been trying to implement for decades with guidelines such as the Surviving Sepsis Campaign and as such they don't address the longstanding issues in sepsis resuscitation. We don't know how much fluid individual patients need, whether fluid responsiveness needs to be corrected and how (norepinephrine alone can correct fluid responsiveness by recruiting blood from the venous system), which patients need vasopressors, when to start vasopressors, what is the correct balance between intravenous fluids and vasopressors, what resuscitation targets should be corrected (what parameters, what targets, are the targets static or dynamic)? Despite decades of research, most of these questions remain unanswered. We intended to test whether the problem could be tackled from a different angle, using reinforcement learning.

However, it is true that 4 hours is quite coarse but it represents an initial trade-off between data availability and the ability to model acute changes. Subsequent iterations of our model may reduce this time interval or use different approaches.

## State, action and transitions

We have discussed above why the WIS estimate of the zero-drug policy cannot be compared to the clinicians' or the AI policies. We would also remind non-clinicians that sometimes some of the hardest things to learn in medicine are when not to give a drug, even though you can.

## Interpretability

We proposed to use the relative feature importance as an indicator of which parameters were important to physicians and the AI clinicians when deciding whether a patient needed to be given a particular drug. Jeter et al. criticised the approach. We defend that the method proposed is appropriate to help elucidate the workings of the AI Clinician compared to the actual clinical actions, even though, we agree that it does not provide interpretability at the level of an individual patient. As Jeter et al. state, it "reveals the top factors contributing to the discrepancy between the clinical policy and the AI clinician policy." This was the intended purpose.

We think that the example provided by the authors (a hypothetical policy "X" withdrawing treatment because of a high comorbidity score) explains very well why this method is appropriate. This is exactly the information that we intend to gain: what parameters lead to a given decision, regardless of whether the decision is to give or not to give a drug.

Next, Jeter et al. question why the two plots (in our original paper's *Supplementary Figure 2*) are dissimilar. They should be dissimilar since the output predicted (whether a drug is given or not) sometimes differ between the two policies, which we can represent using the following table:

| Input features | Drug given by doctors | Drug recommended by AI Clinician |
| --- | --- | --- |
| xx… xx… xx | 0 | 1 |
| xx… xx… xx | 1 | 1 |
| xx… xx… xx | 0 | 0 |
| xx… xx… xx | 1 | 0 |

From a machine learning perspective, we hope it is clear to the reader that the two random forest models should look different, since they are trying to predict different dependent variables. Clearly, one can work towards better ways of explainability in reinforcement learning algorithms, however, this is in itself a nascent field within XAI and also not the main result of our Nature Medicine paper. We much welcome suggestions of better methods to improve the explainability of RL algorithms and actively work ourselves towards it.

# External validation on the eRI cohort

## Sepsis incidence

Jeter et al. wonder which sepsis definition was used in the two databases. To clarify, we did not use different sepsis definitions in the datasets, we used the same. The international sepsis-3 definition was used throughout this analysis, with no modification. The definition was published in 2016 and defines the current state of the art agreed by the intensive care clinical community (Singer M, Deutschman CS, Seymour C, & et al, 2016). We did state that "because of differences between the two datasets, slightly different implementations of the sepsis-3 criteria were used". To be clear, this means that due to differences in the database organisations the data extraction process differed slightly between the two datasets, but not the definition of sepsis. Because of the size of the eRI database (over 3.3 million patients and 2.4 terabytes), once the estimated time of sepsis onset was known, it was not computationally feasible to set the data extraction window individually for every patient, as we did in MIMIC-III. Instead, we only included patients whose sepsis onset was within 36h post ICU admission and extracted their data for the first 72 post ICU admission.

Next, Jeter et al. point out at a supposedly high rate of sepsis in the eRI cohort. The rate of sepsis in this cohort cannot be easily calculated because of subsequent exclusions of subgroups due to insufficient data in this retrospective dataset. This limitation is clearly stated in the original Nature Medicine manuscript. However, we do not understand how Jeter et al. suggest an estimated rate of sepsis of 83.5% in the eRI cohort (which is also not consistent with the incidence rate of sepsis in ICU). The complete patient flow diagram (*Figure J6*) shows that over 900,000 patients were not given antibiotics or had no fluid sampled for microbiology testing. Therefore, these are all ICU patients without sepsis and should be considered in the denominator for calculating rates of sepsis in the eRI cohort. When including only these 901,383 additional patients, the estimated rate of sepsis in this cohort becomes 107,450/1,030,013, or about 10.4%.

## Data quality

The quality of the data in the eRI cohort is brought into question by Jeter et al. The eRI database is a great example of "real world" data: it is large but represents the reality of medical datasets, and as such is messy and imperfect. As an example, we found over 10,000 different free-text labels corresponding to prescriptions of norepinephrine. We took a conservative approach and excluded any hospital where there was doubt about the data quality, which explains why the exclusion rate is so high. Many hospitals collected vital signs, laboratory values and prescriptions but not real-time rate of infusion pumps, and as such were unusable for our project. We established robust and reproducible criteria to exclude hospitals with unsuitable data and documented this in our original paper.

To conclude that excluding these hospitals inherently nullifies the validity of the results due to "poor data quality" is inappropriate and represents a lack of understanding of the data source.

As the eICU system is a supplementary level of care for ICU patients, different ICUs have different implementations of the system. The details of this are explained in the available documentation of the eICU Collaborative Research Database at eicu-crd.mit.edu (Pollard et al., 2018) but we will briefly explain with an example. Hospital laboratory, admission, discharge and transfer (ADT), and bedside vital sign monitoring data are brought into software used by the eICU team with nearly universal coverage via HL-7 interfaces. However, other interfacing of other data may not have universal adoption across hospitals and may vary over time. For example, most hospitals have implemented a pharmacy order interface, directly sharing all validated medication orders with the eICU staff. However, some hospitals were either slow to adopt this technology or have not adopted it due to a variety of institutional reasons such as the lack of IT resources to implement and test this feature.

We are well aware of the potential for inclusion bias but maintain that the method employed - *a priori* exclusion of hospitals who do not exhibit reliable processes to document certain data elements necessary for analysis- is the most robust. It is difficult to imagine how excluding a hospital that does not have IT resources available to test an electronic interface fatally biases the study of individual patient response to fluids and vasopressors. If this were the case, then one would also conclude that all research performed on electronic health record data is fatally flawed due to excluding hospitals using paper charts.

As stated above we do recognise the limitations of retrospective research and include it in our discussion within our manuscript. This is an early scientific paper that establishes proof-of-concept, requires further development and prospective testing, and is not yet a clinically actionable tool.

### External validation

Jeter et al. comment on the different shape of the plots showing the patient mortality as a function of the difference between dose given and dose recommended (*Figure J7*).

First of all, the top figure there is built from the internal validation set, and as such a better performance than in the external validation set is expected. Next, the internal validation set size is much smaller (3,417 versus 80,257 in the eRI cohort), which explains why the plots are noisier and few data points are found at the extremes of the plots.

These differences in doses of fluid and the effect on mortality reflect the uncertainty in existing clinical literature on sepsis resuscitation. It is unclear whether fluid "under-dosing"/restriction (unlike fluid "overdosing"/liberal fluid treatment) is associated with worse outcome in sepsis. For example, the results of the CLASSIC randomized controlled trial in sepsis (Hjortrup et al., 2016) concluded "the patient-centre outcomes all pointed to benefit with fluid restriction…".

Finally, the important inherent limitations of this figure must be remembered. It analyses the average dose gap at the patient level, so that a patient who received 1 L of intravenous fluids in excess, and then had a deficit of 1 L is considered in this analysis to have matched the optimal policy! Looking at the absolute (cumulative) total dose difference at the patient level gives a different picture (*Figure R3*, using a unique model optimising 90-day mortality).

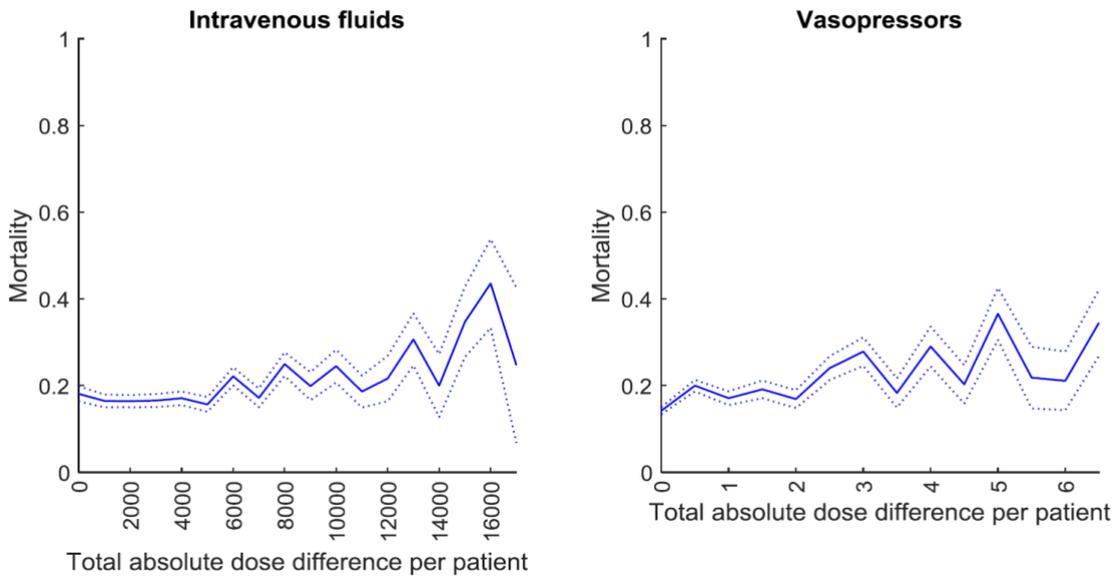

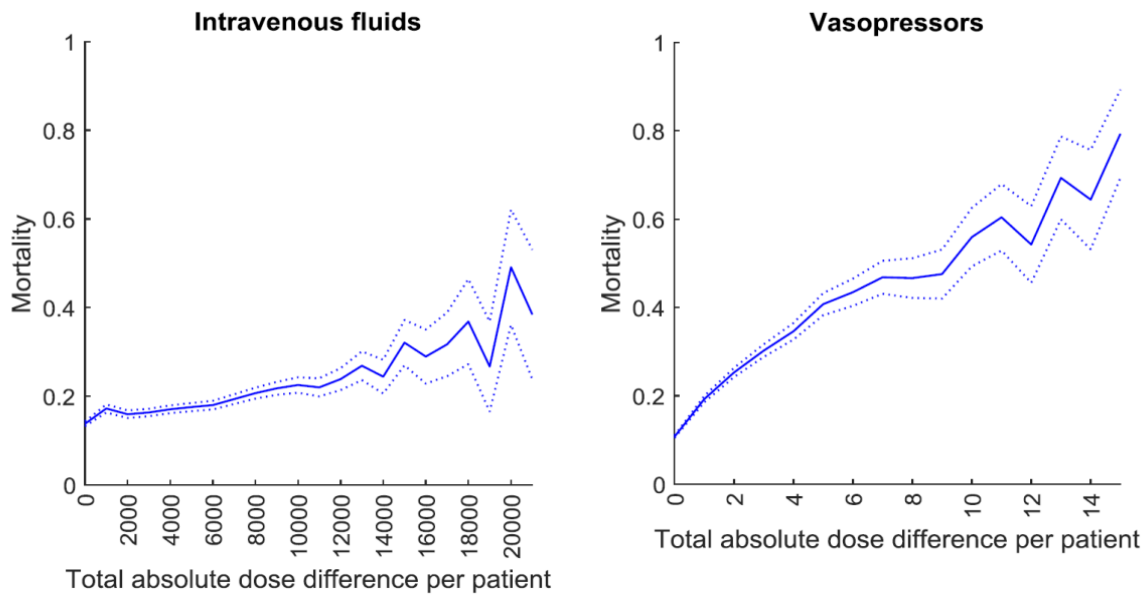

*Figure R3: Total absolute dose difference per patient, for IV fluids and vasopressors, in the MIMIC-III test set (top) and eRI (bottom).*

# Cohort definition

### "Unnecessary AI"

Jeter et al. argue that "nearly two-thirds received no vasopressor treatments in their ICU course, meaning that two-thirds of the patients had no need for the AI Clinician's intervention." This comment implies that human doctors are the gold standard, if no more patients than those who were observed to be on vasopressors should be on vasopressors. We respectfully and strongly disagree with this comment and would like to emphasize that AI algorithms may point towards treatment patterns that are new or infrequent. Vasoplegia is a key feature of sepsis, and an important contributor to hypovolaemia. A higher circulating blood volume can be restored by giving IV fluids and often vasopressors. While it is possible to normalise a patient's blood volume and blood pressure with (sometimes large) volumes of IV fluids, this approach is sometimes harmful, and the dangers of a sustained positive fluid balance in sepsis are well documented (Acheampong & Vincent, 2015; Boyd, Forbes, Nakada, Walley, & Russell, 2011; de Oliveira et al., 2015). In fact, since the publication of our AI Clinician manuscript a double-blind randomised clinical trial has reported that earlier vasopressor therapy compared to standard clinical care led to increased shock control (Permpikul et al., 2019).

The AI clinician suggests a different strategy: that more patients should be on a low dose of vasopressors (looking at *Figure M3c* in the original publication confirms that most patients recommended to receive vasopressors should be on less than 0.22 mcg/kg/min of norepinephrine-equivalent) and recommended to have received less intravenous fluids. As such, it applies to all patients with any severity of sepsis, and not only to patients with shock.

### Iatrogenic treatment

Jeter et al. express concern about the AI Clinician delivering vasopressor to normotensive patients. As detailed above there is certainly equipoise as to how much IV fluid to give, when to start vasopressors and what parameters to target in critically ill patients who have sepsis. There is great variation in everyday clinical practice. Iatrogenic injury can and does arise when the wrong doses of these treatments are given. The AI Clinician has been designed to "learn" to optimise these decisions. It too could have learnt some incorrect decisions in this first model. But this is why we state that the tool is not yet ready for clinical use and requires further development and prospective testing.

## Final comments

Like Jeter et al., let us also end with a quote, this one borrowed from George Box, who said that "all models are wrong, but some are useful." We acknowledge that all science has limitation and we have been fully transparent about the limitations of our model. We do share the authors' appreciation of evidence-based medicine, which we emphasized when we concluded that "this work will clearly require prospective evaluation using real-time data and decision-making in clinical trials".

We foresee that it will be the unification of two very different schools of reasoning: medical statistics (causal modelling) and artificial intelligence (reinforcement learning) that will ultimately yield the largest benefits in healthcare. The novelty to medicine is that our reinforcement learning approach demonstrates that growing our reasoning horizon beyond single decisions, as in classical causal inference, to the scale of overall treatment strategies, bears the promise of significant advances in healthcare – since patient care is often more than the sum of its parts.

## Competing interests

The authors declare the following competing interests: M.K. does not have competing financial interests. L.A.C. receives funding from Philips Healthcare. O.B. is an employee of Philips

Healthcare. A.C.G. reports that outside of this work he has received speaker fees from Orion Corporation Orion Pharma and Amomed Pharma. He has consulted for Ferring Pharmaceuticals, Tenax Therapeutics, Baxter Healthcare, Bristol-Myers Squibb and GSK, and received grant support from Orion Corporation Orion Pharma, Tenax Therapeutics and HCA International with funds paid to his institution. A.A.F. has received funding from Fresenius-KABI. The financial interests have not changed since the Nature Medicine paper's publication.